# Factored Latent Analysis for far-field tracking data

Chris Stauffer
Computer Science and Artificial Intelligence Laboratory
Massachusetts Institute of Technology
Cambridge, MA 02148

## Abstract

This paper uses Factored Latent Analysis (FLA) to learn a factorized, segmental representation for observations of tracked objects over time. Factored Latent Analysis is latent class analysis in which the observation space is subdivided and each aspect of the original space is represented by a separate latent class model. One could simply treat these factors as completely independent and ignore their interdependencies or one could concatenate them together and attempt to learn latent class structure for the complete observation space. Alternatively, FLA allows the interdependencies to be exploited in estimating an effective model, which is also capable of representing a factored latent state. In this paper, FLA is used to learn a set of factored latent classes to represent different modalities of observations of tracked objects. Different characteristics of the state of tracked objects are each represented by separate latent class models, including normalized size, normalized speed, normalized direction, and position. This model also enables effective temporal segmentation of these sequences. This method is data-driven, unsupervised using only pairwise observation statistics. This data-driven and unsupervised activity classification technique exhibits good performance in multiple challenging environments.

## 1 INTRODUCTION

A human describing the activity in an environment will call on generalized concepts to describe each of a number of characteristics of the object and its activity. For instance, a person walking west on the school's front sidewalk or a car stopping at the second toll both are very compact descriptions that imply particular sizes, velocities, directions, and locations. Using this low-bandwidth description, it is possible to describe activity sequences in a compact manner. For instance, the person walked away from the car park, stopped by the loading dock, and ran back towards the car park.

An effective description of large numbers of complete tracking sequences should be both factored and segmental. With a **factored** representation, one can describe whether a person is stopped, walking, or running independently of the direction they are moving, or one can describe the lane of traffic independently of the vehicle type. With a **segmental** representation, one can describe a person that runs to the ATM, stops for two minutes, and walks off towards the east.

Most previous work in *learning* effective descriptions of tracked objects did not employ a factored model. Different aspects of the description were either treated completely independently or thrown into the same feature space. The first approach doesn't exploit the information in the joint observations. The second approach uses the joint observations but doesn't provide an effective description of the individual aspects of the description.

Also, previous work in *learning* descriptions of tracked object sequences is often not segmental. Many approaches are only capable of describing the entire sequence, not each individual action in the sequence. For most tracking sequences in interesting environments, a single description is not sufficient for the entire sequence as the object may change direction, shape, speed, or move from one region to another.

The goal of this work is to use large sets of observations of object state over time to automatically generate an unsupervised description of the activity of the objects in the scene solely from pairwise observation statistics. This system is capable of compactly describing in what state an object is and what it is doing throughout an entire tracking sequence. With very little supervision, these latent classes can be associated with certain labels associated with high-level human concepts. Building such an intermediate representation that accurately represents the observations for a particular environment is extremely useful for unsupervised and semi-supervised description.

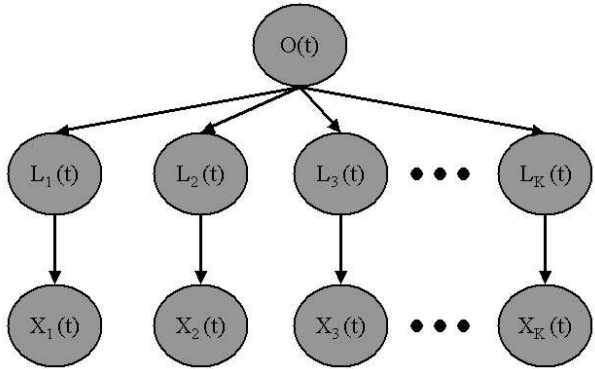

Figure 1: The graphical model used in Factored Latent Analysis (FLA).

### 1.1 PREVIOUS WORK

Our primary goal is to determine a set of latent class models for each of a number of different observable characteristics. Graphical models including latent class models have become increasingly popular over the last few years. Thomas Hofmann's aspect model[2] is capable of inducing latent word and document models from occurrence statistics of words within documents. Latent Dirichlet Allocation[1] adds the ability to represent the process by which documents are drawn as *mixtures* of latent word classes as determined by a Dirichlet prior.

Our model and estimation is different in two respects. First, we explicitly factor the observations as shown in Figure 1. This results in a separate latent class model for each of a number of different descriptions allowing direction, speed, size, and location to have completely independent descriptions. Second, we estimate the model using pairwise observation statistics. Stauffer [6] built hierarchical models using pairwise joint co-occurrence of observations of a single type. This work is similar, but uses all the pairwise observation between different characteristics to induce latent class models that are consistent with those statistics across multiple aspects.

Many varieties of data-driven perceptual data mining[5] techniques have been applied to activity analysis. Johnson and Hogg [3] clustered trajectories in a scene into 400 representative clusters allowing generalization and prediction, but not compact description. Stauffer and Grimson [6] used a hierarchal clustering technique to develop a binary tree classification hierarchy, which is somewhat more compact but also not segmental. Because these approaches describe entire paths holistically it is unclear how they can be adapted to describe compound activities. Makris and Ellis[4] also clustered entire trajectories into independent paths, but they extracted common features of paths to describe way points enabling some segmentation of the entire sequence but still a limited description of each segment.

This paper describes a method for learning an effective factored segmental description of activity in scenes directly from the tracking data using observed pairwise joint co-occurrences. Section 2 discusses the statistical model and the standard estimation technique. Section 3 describe an alternative approximate technique that remains computationally tractable as the number of observations sequences becomes large. Section 4 shows results from multiple environments. Section 5 and Section 6 discuss future work and conclusions of this work.

## 2 FACTORED LATENT ANALYSIS

The only aspects of the state of an object that can be easily modelled are those aspects that are directly reflected in the observations of an object, such as height, size, shape, speed, direction, location or color. This section describes our generative model of observed characteristics of an object.

At time $t$, the object state is $O(t)$ and the observed characteristics of the object are $X(t)$. $X(t)$ is a vector of observations $\{X_1(t), X_2(t), ..., X_K(t)\}$. Our goal is to compactly represent

$$p(X(t), O(t)) = p(X(t)|O(t))p(O(t)) \qquad (1)$$

where $p(O(t))$ is the probably of being in a particular state and $p(X(t)|O(t))$ is the conditional likelihood of producing a particular set of observations.

Figure 1 shows the graphical model used in Factored Latent Analysis. In the model, the observed state $X(t)$ is factored into $(X_1(t), X_2(t), ..., X_K(t))$ and each separate type of observation is represented by a compound latent state $L(t)$, which is factored similarly into separate latent classes $(L_1(t), L_2(t), ..., L_K(t))$. Thus, the latent class of a particular object at a particular time is represented by a $K$-tuple of latent state assignments $(l_1, l_2, ..., l_K)$. We assume that the observation $X_i(t)$ can be drawn independently from a distribution characterized by one of a small number of latent classes for that observation type at that time, or $L_i(t)$. This is the latent class conditional output distribution $p(X_i|L_i)$.

For instance, observations of normalized size $X_s(t)$ ideally would be a single value with additive noise. In an environment with people, cars, and trucks, a coarse approximation would be three latent size classes, each exhibiting a particular size with additive measurement noise. In many surveillance environments, the object's speed and movement direction at a particular time can be characterized by a particular model.

Our approximation to $p(O(t))$ is a prior likelihood of every possible combination of latent class label assignments $p(l_1, l_2, ..., l_K)$ and latent class conditional output distributions $p(x_i|l_i)$. Thus, based on the independence assumptions in this model, equation 1 becomes

$$p(x_1, x_2, ..., x_n) = \sum_{(l_1, l_2, ...l_n) \in \mathbf{L}} p(l_1, l_2, ..., l_K) \prod_{i=1}^{K} p(x_i|l_i), \quad (2)$$

where $\mathbf{L}$ is the set of all possible combinations of latent aspect assignments. Of course, many of those combinations could have a negligible probability of occurring. The number of potential combinations is

$$|\mathbf{L}| = \prod_{i=1}^{K} k_i, \quad (3)$$

where $k_i$ is the number of latent classes for each type of observations. The number of latent classes for observation type $i$, $k_i$, is the primary means by which the capacity of this model is controlled. In the examples in this work, this number has been chosen, but future work will address model selection for FLA.

### 2.1 EM ESTIMATION

Given the number of latent classes and a large set of observations, it is possible to estimate the maximum likelihood parameters of the model. The remainder of this section describes how the entire joint latent model could be estimated directly from raw data. The potential advantages and pitfalls of this technique are outlined. Due to the complexity of the model and the quantity of data in the domain we are considering, a computationally feasible alternative is introduced in Section 3. This alternative is an approximate technique for estimating the model from only aggregate pairwise observation statistics between different observation types.

The obvious approach to estimating the parameters of the model is a "simple" EM implementation. The hidden variable is the latent class assignment. The E-step involves computing the likelihood over each latent variable given model of $p(l_1, ..., l_K)$ and $p(x_i|l_i)$. The M-step involves maximizing the model parameters given the latent likelihoods. The exact form of the model will dictate the complexity of both steps. For instance, $p(x_i|l_i)$ could be a multinomial distribution, a Gaussian distribution, or a mixture of Gaussians and $p(l_1, ..., l_K)$ could be a multinomial joint distribution with a uniform prior, a Dirichlet prior, or any other alternative.

Using this approach, one can estimate the full joint distribution $p(l_1, l_2, ..., l_K)$ over the latent classes using all the data while simultaneously estimating the class conditional output distributions $p(x_i|l_i)$. Unfortunately, the computational complexity of this procedure is $O(Nk_1k_2...k_K)$, where $N$ is the number of observation vectors observed. For a nominal number of latent classes (approximately 3 or 4) over a nominal number of observation types (3 or 4) over the number of observations in a single hour of tracking, a single EM iteration can take as much as a day on current hardware. For this reason, we present the alternative in the next section.

## 3 ESTIMATION FROM PAIRWISE OBSERVATION STATISTICS

This section describes our parameter estimation technique for the Factored Latent Analysis model as applied to the problem of activity analysis. Four steps are described: quantization of the (potentially) continuous output observations; estimation of pairwise joint observation statistics; estimating the latent class likelihoods and mixing probabilities from the pairwise statistics; and classification using the model.

### 3.1 QUANTIZATION OF OBSERVATION STATISTICS

Our first approximation is to discretize the output observations for of each type of observation. This can be done by uniformly binning the output space or using vector quantization to more compactly represent the output space. Thus, potentially continuous observations, $x_i$, can be represented by a discrete set of prototypical values $\dot{x}_i$.

Figure 2 shows a scene with overlayed tracking data. For this scene, we will make discretized measurements of size, velocity, direction, and position. In most cases, after normalizing the tracking measurements using an automated data-driven technique developed by Stauffer et al. [7], we can simply partition the measurements from their min to max value into equal sized bins. Using 64 bins to represent sizes, velocities, and directions is sufficient, whereas 2D position requires more bins.

In theory, this quantization is not necessary, but it makes it computationally feasible to collect estimates of joint occurrences of observations of type $i$ and type $j$, or $\hat{p}(x_i, x_j)$. The primary advantage is that the discretized joint occurrence estimate $\hat{p}(\dot{x}_i, \dot{x}_j)$ will remain a constant size regardless of the number of observations. Because there are potentially tens of thousands of tracked objects per hour and each tracked object can contain hundreds of valid windows, this approximation is require to make this application computationally feasible.

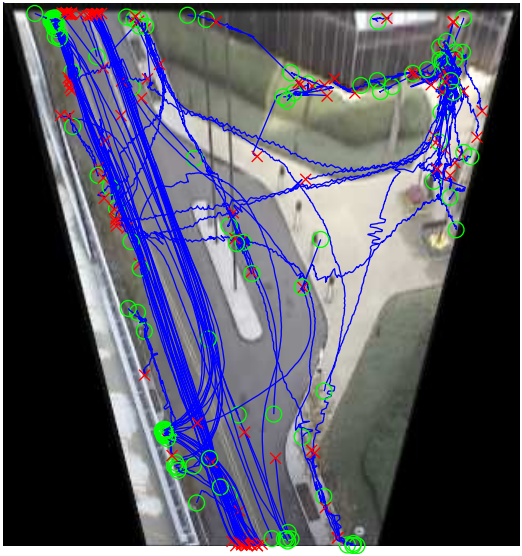

Figure 2: This figure shows the traffic in a particular environment after rectification. Green circles denote beginnings of track and red x's denote ends of tracks. A large portion of the traffic is on the road moving either north(up) or south(down), but pedestrians are often seen crossing the road and moving to or away from the office building on the right.

### 3.2 ESTIMATING PAIRWISE JOINT OBSERVATIONS

If one was provided segmented, labeled data for a particular observation type, the class conditional probabilities $p(x_i|l_i)$ could be estimated directly. Unfortunately, our data is neither segmented or labeled for any of the types of observations.

Because we lack an oracle to segment the sequences for each observation type, we make an assumption that the observations within a Gaussian-weighted temporal window are drawn from the same latent class. This assumption is an integral part of this learning technique. For object activity, we use a window of one second duration. Within such a window an object's size, velocity, direction of travel and location are assumed to be drawn independently from a single latent class for each observation type.

For instance, over a single second a particular object could be described as a small, fast-moving object travelling north on the highway. This corresponds to one possible N-tuple joint latent state assignment in the set **L**. If the temporal window size is too small, a randomly chosen window would exhibit little within-class variability in the observations. If the temporal window is too large, many randomly sampled windows would include observations drawn from multiple latent classes.

$\hat{p}(\dot{x}_i, \dot{x}_j)$, is an estimate of the likelihood that a pair of observations would be drawn from the same underlying joint latent class. Given a randomly sampled window of observations and a weight mask $m(dt)$, $\hat{p}_n(\dot{x}_i, \dot{x}_j)$ is a frequentist approximation to the probability of drawing two observations independently from the window.

This estimate can be accumulated in an online fashion, because

$$\hat{p}(\dot{x}_i, \dot{x}_j) = \sum_{n=1}^{N} w(n) p_n(\dot{x}_i, \dot{x}_j), \quad (4)$$

where $p_n(\dot{x}_i, \dot{x}_j)$ is the joint cooccurrence within a particular window and $w(n)$ is a weight for that window. Thus, the aggregate estimate and the sum of the weights are sufficient statistics. In this work the weight was assumed to be uniform, but the weight can be used to decrease the affect of extremely redundant observed states. For instance, the weight for the hundreds of redundant observations of a stopped car can be decreased to lessen their total effect on $\hat{p}(\dot{x}_i, \dot{x}_j)$.

Figure 3 shows $\hat{p}(\dot{x}_s, \dot{x}_v)$ for a scene containing pedestrians and vehicles, where $x_s$ is an observation of object size and $x_v$ is an observation of object velocity. In this case, "size" is estimated as the square root of the number of pixels and "velocity" refers to the speed an object is travelling in normalized coordinates after rectification. As is evident in the figure, there are two main types of objects: slow-moving small objects and fast-moving large objects. Although, it is interesting to note that both the small objects and the large objects can be observed in a stopped or loitering state.

### 3.3 ESTIMATING PAIRWISE LATENT CLASS CONDITIONAL AND PRIOR DISTRIBUTIONS

Given the estimate of $\hat{p}(\dot{x}_s, \dot{x}_v)$ for pairs of observations of size and velocity and a number of latent size classes $k_s$ and number of latent velocity classes $k_v$, it is possible to estimate the model parameters that best approximate those joint statistics. This is an adaptation of PLSA [2] or similar latent class models to joint observations rather than equivalence sets of observations. I.e., we are estimating the model from pairs of observations (e.g., word-image pairs) that appear in the same equivalence set rather than sets of observations (e.g., document-word pairs).

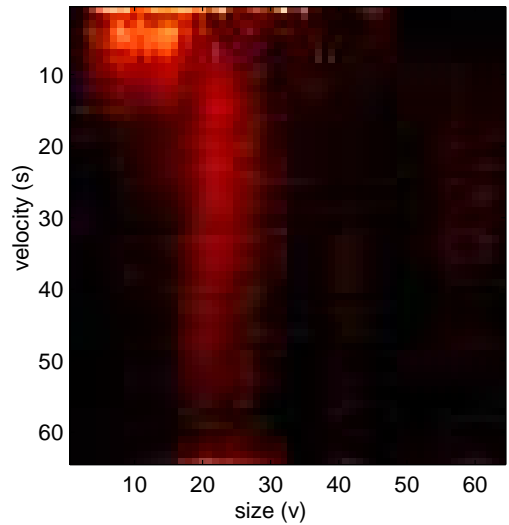

Figure 3: An example joint occurrence over size and velocity, or $\hat{p}(\dot{x}_s, \dot{x}_v)$, estimated from a scene containing pedestrians and vehicles. Brighter indicate higher likelihood of cooccurrences within a window of observations drawn from a single joint latent class.

Given the model's estimates of the mixing likelihoods $\tilde{p}(l_s, l_v)$, and conditional output likelihoods for $\tilde{p}(x_s|l_s)$ and $\tilde{p}(x_v|l_v)$ the likelihood of observing a particular pair of observations from the model is simply

$$\tilde{p}(\dot{x}_i, \dot{x}_j) = \sum_{(l_s, l_v) \in \mathbf{L_{sv}}} \tilde{p}(l_s, l_v) \tilde{p}(x_s|l_s) \tilde{p}(x_v|l_v), \quad (5)$$

where $\mathbf{L_{sv}}$ is the set of all pair of latent size and velocity assignments.

Given random initial estimates of $\tilde{p}(l_s, l_v)$, $\tilde{p}(x_s|l_s)$, and $p(x_v|l_v)$, the model parameters that maximize the likelihood of the data can be iteratively estimated using an EM procedure as shown in [2]. This equates to minimizing the KL divergence between the joint statistics of the model $\tilde{p}(\dot{x}_s, \dot{x}_v)$ and the observed joint statistics $\hat{p}(\dot{x}_s, \dot{x}_v)$.

Figure 4 shows the maximum likelihood model for the $\hat{p}(\dot{x}_s, \dot{x}_v)$ example shown in Figure 3 after convergence. It is evident by comparing the two joints, that the model is able to effectively approximate the joint statistics with only two latent size classes and three latent velocity classes. The two latent size classes correspond to pedestrians and vehicles. The bimodal pedestrian class results from individual pedestrians and pairs of connected pedestrians. The three latent velocity classes roughly correspond to loitering, walking, and driving through the scene. As is evident in the mixing matrix, $\tilde{p}(l_s, l_v)$, on the upper left, most pedestrians walk, most vehicles drive, but both classes are sometimes observed loitering for periods of time.

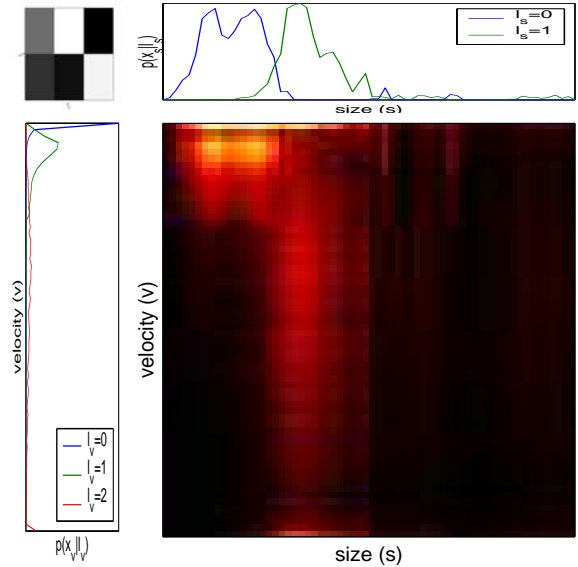

Figure 4: This figure shows: the latent pair likelihoods, $\tilde{p}(l_s, l_v)$, on the upper left; the conditional likelihood models for each latent object size class, $\tilde{p}(x_s|l_s)$, on the upper right; the conditional likelihood for each latent object velocity class, $\tilde{p}(x_v|l_v)$, on the lower left; and the models implied joint statistics as outlined in Equation 5 on the lower right.

While this process could be done independently for each possible pair of observations, it would result in $K$ sets of latent class models for each independent factor. E.g., 2 latent class models for size given velocity; 2 latent class models for size given direction; and so on. The next subsection describes how an estimate of only $K$ sets of latent models can be estimated that are consistent with $K^2$ pairwise joint estimates.

### 3.4 EXPLOITING $K^2$ PAIRWISE JOINT OBSERVATIONS

Factored Latent Analysis includes $K$ sets of latent classes $\tilde{p}(\dot{x}_i|l_i)$, one for each type of latent observation, and a model for drawing latent class combinations $\tilde{p}(l_i, l_j)$. The goal of our approximation is to estimate the K sets of latent class models for each independent factor and the mixing proportions that best approximate *all* the pairwise observations. One can trivially estimate $K^2$ pairwise joint observation estimates[1], each of which corresponds to a possible pair of observation types.

---

[1] Since $\hat{p}(\dot{x}_i, \dot{x}_j)$ is redundant with $\hat{p}(\dot{x}_j, \dot{x}_i)$, only $\frac{K(K+1)}{2}$ pairs need to be stored.

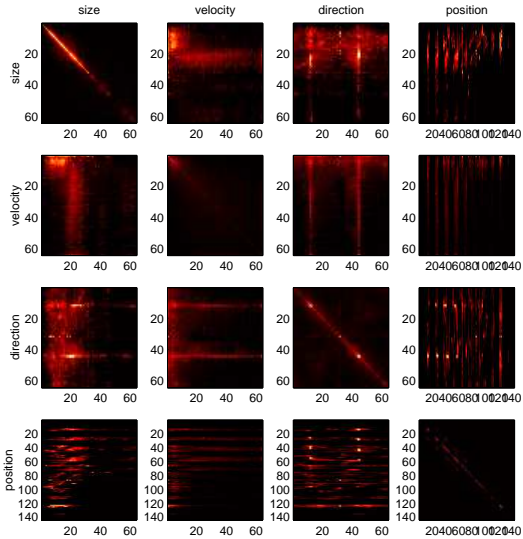

Figure 5: This figure shows 16 matrices corresponding to each possible $\hat{p}(\dot{x}_i, \dot{x}_j)$ for the scene shown in Figure 2 given a one second equivalency window.

Figure 5 shows all pairwise joint observations for the scene shown in Figure 2. It is obvious that there is significant structure in these pairwise joint statistics. For instance, small objects tend to loiter or move slowly. Large objects tend to move faster in one of two primary directions (north and south). Because of the 2D nature of position, it is difficult to interpret the last row and column without performing classification.

Given a complete set of $K$ latent class conditional output distributions $\tilde{p}(\dot{x}_i|l_i)$ and $K^2$ pairwise latent marginal distributions $\tilde{p}(l_i, l_j)$, every marginal pairwise output distribution $\tilde{p}(\dot{x}_i, \dot{x}_j)$ can be estimated. By minimizing the KL divergence between all of the pairwise output distributions and pairwise output estimates, the likelihood of the data given the model can be maximized. The criterion is

$$\tilde{\theta}^* = \underset{\tilde{\theta}}{argmin} \sum_{i,j} d(\hat{p}(\dot{x}_i, \dot{x}_j)||\tilde{p}(\dot{x}_i, \dot{x}_j)), \quad (6)$$

where $\tilde{\theta}$ is the parameters of the $K$ sets of $k_i$ multinomials over $|\dot{x}_i|$ and the $K^2$ $k_i$ by $k_j$ latent joint distributions. This equates to maximizing the likelihood of *all* pairwise measurements.

Figure 6(a) shows the model parameters $\tilde{\theta}$ and Figure 6(b) shows the corresponding $\tilde{p}(\dot{x}_i, \dot{x}_j)$ for the same scene after convergence. This model was automatically estimated given two, three, six, and eight latent classes respectively for size, velocity, direction, and position.

The latent class conditional distributions have intuitive interpretations. As in the previous subsection, the latent size classes roughly correspond to pedestri-ans and vehicles even though the distribution of pedestrian sizes is bimodal. This is due to the fact that pedestrians and pairs of pedestrians exhibit similar speeds, directions, and locations. The same velocity classes also occur. Using only the diagonal elements $\tilde{p}(\dot{x}_i, \dot{x}_i)$ of Figure 6(a), the likelihood of each latent class is evident. There are two common directions corresponding to the road directions. These directions are taken by both pedestrians and vehicles, but there are many directions that are taken primarily by pedestrians.

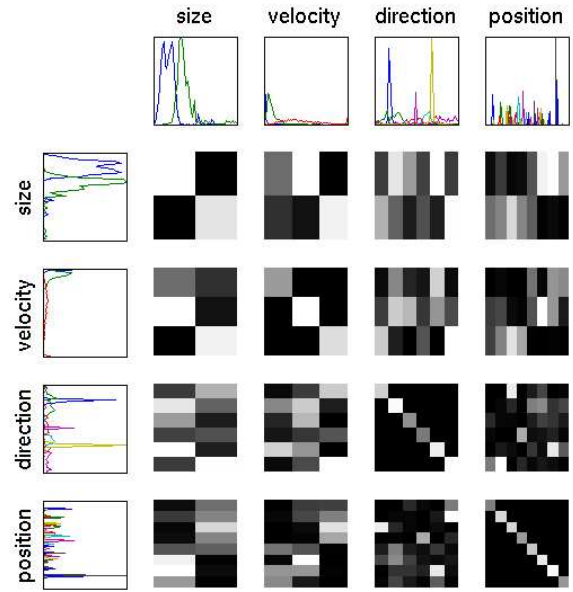

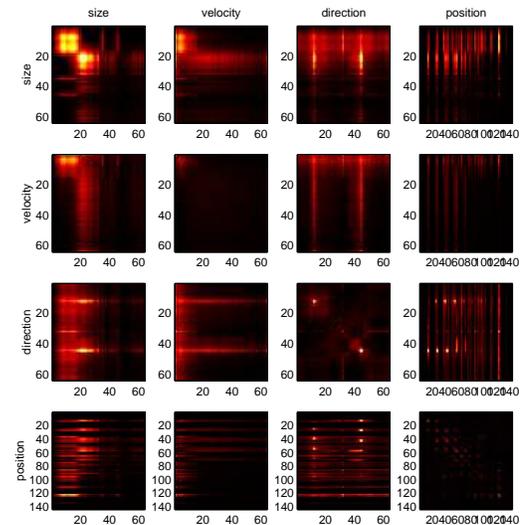

Figure 6: This figure shows (a) the latent classes and their mixing ratios and (b) the resulting $\tilde{p}(\dot{x}_i, \dot{x}_j)$.

### 3.5 APPLYING THE MODEL TO CLASSIFICATION

While any state observation $X(t)$ could be classified into one of each type of latent class independently, it is important to remember what the class conditional likelihoods actually represent. They were calculated based on an assumption of temporal coherency, thus the latent class likelihoods for a particular time $t$ are estimated from more than simply $X(t)$.

Currently, we simply assume the latent class posteriors are the weighted expectation of the independent posterior estimates over the observations temporal coherency window. This performs smoothing which greatly reduces the number of spurious detections from instantaneous tracking glitches caused by lighting effects, occlusions, and various tracking abnormalities.

## 4 RESULTS

Figure 7(a) shows a new set of tracking data from the previous scene. The latent classes from the unsupervised training are nearly identical to the example in the previous section despite differences in the number of objects or distributions of their behaviors. Figure 7 (b)-(e) show the maximum likelihood classification of the observation for hundreds of sequences. See the figure for a detailed description. These results illustrate how position latent classes tend to represent areas which contain similar types of object moving in similar ways. In Figure 7(e) the blue class corresponds to an area where cars and pedestrians are present and both are likely to be in the "loitering" state.

Figure 8 show results for a different environment with the same number of latent classes. Both examples estimated a set of latent classes for all four types of observations that are specific to a particular environment and input.

By a simple process of labeling these latent classes with textual descriptions, interesting compound queries can be constructed. E.g., "show me the pedestrians that crossed the road." "How many vehicles stopped in the loading zone?" "How many individuals who stopped in the scene started in the opposite direction?" "How much of the traffic on the road is vehicles?" "Did anything stop in the scene for more than 30 seconds?" These queries illustrate the importance of factored representation and the ability to segment effectively.

## 5 FUTURE WORK

We look forward to implementing an efficient estimation technique for estimation of the complete $p(l_1, l_2, ..., l_K)$ as described in Section 2. This would have a number of advantages. Rather than having estimates of all pairwise latent joints $\tilde{p}(l_i, l_j)$, we would have the full latent joint distribution, $\tilde{p}(l_1, l_2, ..., l_K)$. This is obviously a more desirable approach but is currently unimplementable for our domain of interest.

Another major issue is model selection, i.e. choosing the proper number of latent classes for each observation type. We intend to implement a model selection criteria that will incorporate a measure of the mutual information in the latent joint distribution. This will make splitting a latent class for one type of observation undesirable if it does not increase the information about other types of observations. For instance, it will be more desirable to split vehicles into cars and trucks if they exhibit different speed, direction, or location characteristics. Related to the issue of choosing the number of classes is learning class labels. In fact, labels given from an oracle could simply be another observation type.

Finally, there are many more characteristics that can be included in this estimation. This could include average color, color histograms, shape, sparsity, internal object deformation, source location, or sink location. Each of these characteristics may bear new information on object activity, class, or appearance.

## 6 SUMMARY

We have presented a method for approximating the Factored Latent Analysis model using pairwise joint observation estimates $\hat{p}(\dot{x}_i, \dot{x}_j)$. This factored model enables a compact description of object properties, which is more useful than many previous automatically-generated unfactored descriptions. Our model also enables tracking sequences to be segmented into component parts effectively. This is essential in describing compound activities that occur in most environments.

This algorithm exhibited good performance across many environments by exploiting the pairwise observations and temporal coherence. We believe this work

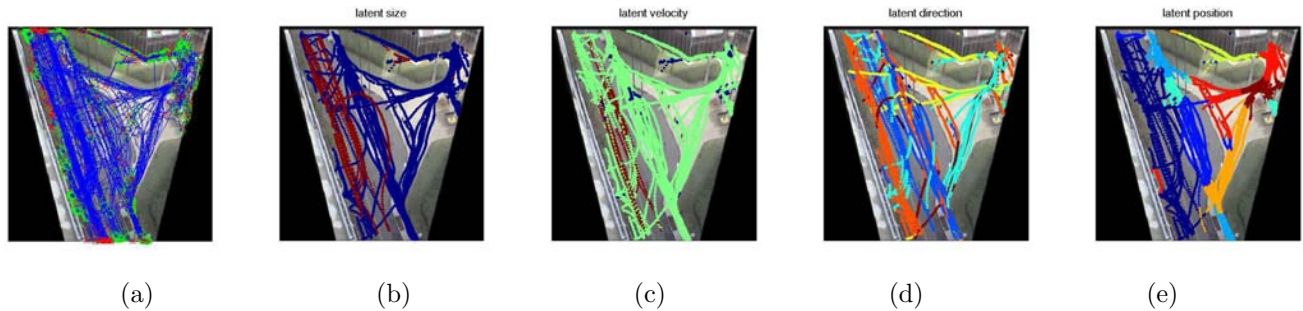

(a)   (b)   (c)   (d)   (e)

Figure 7: This figure shows a new set of tracked objects for the previously introduced scene as well as the classification of individual object states in each of the $K$ latent class types. The latent size classes show that vehicles tend to be on the road or in the drop-off region. The latent velocity classes show that only the vehicles passing through on the north-south road travel faster than the nominal speed. Interestingly, there are some examples of pedestrian sized objects going the faster latent velocity class in the road region (bicycles and rollerbladers). Figure 7(d) shows north (orange) and south (blue) classes as well as traffic moving towards the office from the south (cyan) and from the north (yellow) and traffic moving away from the office (red). Figure 7(e) shows regions with similar characteristics. The road (dark blue) is a region unto itself. The sidewalk along the street on the north and south side of the scene (light blue) is split by either walking through the drop-off zone (blue) or through the angled sidewalks (orange and red).

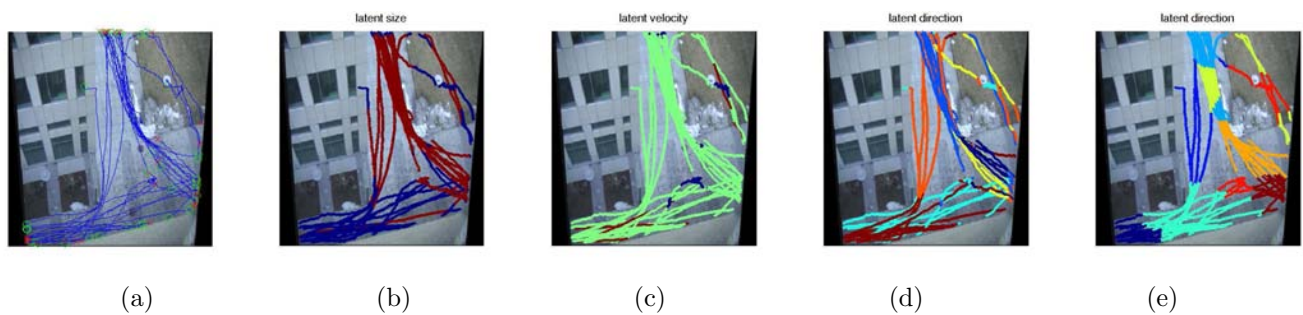

(a)   (b)   (c)   (d)   (e)

Figure 8: This figure shows tracked data for pedestrians in a courtyard (a) and each observation classified in each of the $K$ latent class models. The latent sizes primarily result from occluded and unoccluded pedestrians. Three pedestrians jogged away from the building and four stopped. The stopping zones were clustered together (red) as were the major lanes of movement. There are north, south, east, west, northwest and southeast directions.

shows promise and intend to pursue further related avenues of research.